\documentclass[runningheads]{llncs}

\usepackage{amsmath,amsfonts,bm}

\def\eqref#1{equation~\ref{#1}}

\def\1{\bm{1}}

\newcommand{\test}{\mathcal{D_{\mathrm{test}}}}

\def\vc{{\bm{c}}}

\def\vp{{\bm{p}}}

\def\vt{{\bm{t}}}

\def\vv{{\bm{v}}}

\def\vx{{\bm{x}}}

\def\mC{{\bm{C}}}

\def\mI{{\bm{I}}}

\DeclareMathAlphabet{\mathsfit}{\encodingdefault}{\sfdefault}{m}{sl}
\SetMathAlphabet{\mathsfit}{bold}{\encodingdefault}{\sfdefault}{bx}{n}

\def\sR{{\mathbb{R}}}

\usepackage{graphicx}

\usepackage{tikz}
\usepackage{comment}
\usepackage{amsmath,amssymb} 
\usepackage{color}
\usepackage{hyperref}
\usepackage[T1]{fontenc}
\usepackage{booktabs}
\usepackage{multirow, tabularx, caption}
\usepackage{pifont}

\usepackage[accsupp]{axessibility}  

\begin{document}
\pagestyle{headings}
\mainmatter
\def\ECCVSubNumber{4422}  

\title{Contextual Text Block Detection \\ towards Scene Text Understanding} 


\titlerunning{Contextual Text Block Detection}
%
\author{Chuhui Xue\inst{1,2}\and
Jiaxing Huang\inst{1}\and
Shijian Lu\inst{1}\and
Changhu Wang\inst{2}\and
Song Bai\inst{2}}
\authorrunning{Xue et al.}
%
%
\institute{Nanyang Technological University
\and
ByteDance Inc.\\
\email{xuec0003@e.ntu.edu.sg}, \email{\{jiaxing.huang,shijian.lu\}@ntu.edu.sg},
\email{\{changhu.wang,songbai.site\}@gmail.com}}
\maketitle
\begin{abstract}
Most existing scene text detectors focus on detecting characters or words that only capture partial text messages due to missing contextual information. For a better understanding of text in scenes, it is more desired to detect contextual text blocks (CTBs) which consist of one or multiple integral text units (e.g., characters, words, or phrases) in natural reading order and transmit certain complete text messages. This paper presents contextual text detection, a new setup that detects CTBs for better understanding of texts in scenes. We formulate the new setup by a dual detection task which first detects integral text units and then groups them into a CTB. To this end, we design a novel scene text clustering technique that treats integral text units as tokens and groups them (belonging to the same CTB) into an ordered token sequence. In addition, we create two datasets SCUT-CTW-Context and ReCTS-Context to facilitate future research, where each CTB is well annotated by an ordered sequence of integral text units. Further, we introduce three metrics that measure contextual text detection in local accuracy, continuity, and global accuracy. Extensive experiments show that our method accurately detects CTBs which effectively facilitates downstream tasks such as text classification and translation. The project is available at \url{https://sg-vilab.github.io/publication/xue2022contextual/}. 
\keywords{Scene Text Detection}
\end{abstract}

\begin{figure}[!t]
  \centering
  \includegraphics[width=\linewidth]{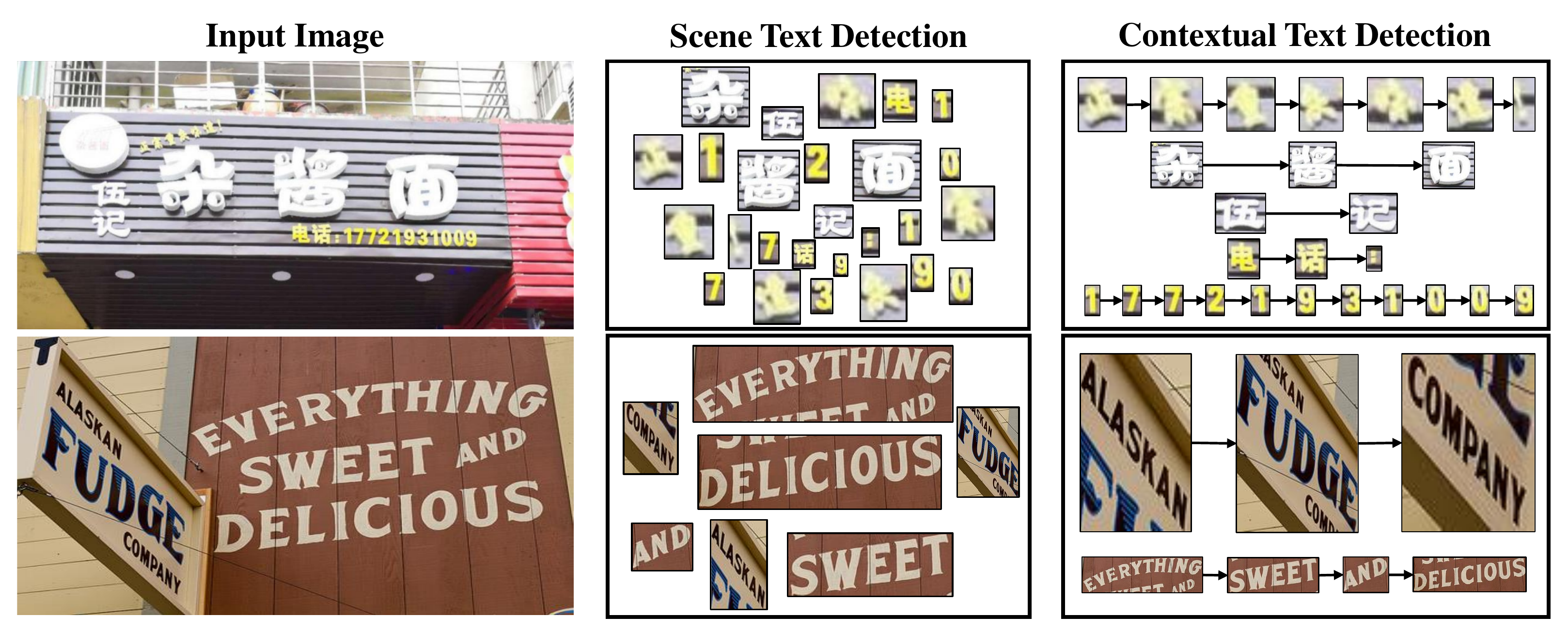}
\caption{
\textbf{Illustration of traditional scene text detection and the proposed contextual text detection:} Traditional scene text detection detects integral text units (e.g., characters or words as shown in the second column) which usually deliver incomplete text messages and have large gaps towards scene text understanding. In contrast, the proposed contextual text detection detects contextual text blocks each of which consists of multiple integral text units in natural reading order. It facilitates the ensuing tasks in natural language processing and scene understanding greatly.
}
\label{fig:intro_samples}
\end{figure}

\section{Introduction}

Scene texts often convey precise and rich semantic information that is very useful to visual recognition and scene understanding tasks. To facilitate reading and understanding by humans, they are usually designed and placed in the form of 
{\fontfamily{pcr}\selectfont contextual text blocks} which consist of one or multiple {\fontfamily{pcr}\selectfont integral text units} (e.g., a character, word, or phrase) that are arranged in natural reading order. Contextual text blocks deliver complete and meaningful text messages and detecting them is critical to the ensuing tasks such as natural language processing and scene image understanding. 

Most existing scene text detectors \cite{long2021scene,liao2020mask,zhang2021scene,liao2020real} focus on detecting integral text units only as illustrated in the second column of Fig. \ref{fig:intro_samples}. Their detection thus cannot convey complete text messages, largely because of two factors. First, they capture little contextual information, i.e., they have no idea which text units are from the same sentence and deliver a complete message. Second, they capture little text order information, i.e., they have no idea which is the previous or the next text unit in the natural reading order. Without contextual and text order information, the outputs of existing scene text detectors have a large gap towards natural understanding of scene texts and relevant scene images.

We propose a new text detection setup, namely contextual text detection, where the objective is to detect {\fontfamily{pcr}\selectfont contextual text blocks} (consisting of one or multiple ordered {\fontfamily{pcr}\selectfont integral text units}) instead of individual integral text units. This new setup has two challenges. First, it needs to detect and group the integral text units into a contextual text block that transmits a complete text message. Several studies~\cite{tian2015text,tian2016detecting} adopt a bottom-up approach by first detecting characters (or words) and then grouping them into a word (or a text line). However, their detected texts usually deliver partial text messages only, e.g., one text line in a contextual text block consisting of multiple text lines in Fig. \ref{fig:intro_samples}. Second, it needs to order the detected integral text units belonging to the same contextual text block according to the natural reading order. Though some work~\cite{li2020end} studies text sequencing in document images, it assumes a single block of text in document images and cannot handle scene images which often have multiple contextual text blocks with very different layouts and semantics. 

We design a \underline{C}ontext\underline{u}al \underline{T}ext D\underline{e}tector (CUTE) to tackle the contextual text detection problem. CUTE models the grouping and ordering of integral text units from a NLP perspective. Given a scene text image, it extracts contextual visual features (capturing spatial adjacency and spatial orderliness of integral text units) of all detected text units,  transform the features into feature embeddings to produce integral text tokens, and finally predicts contextual text blocks. In addition, we create two new datasets ReCTS-Context and SCUT-CTW-Context where each contextual text block is well annotated as illustrated in Fig. \ref{fig:intro_samples}. For evaluation of contextual text detection, we also introduce three evaluation metrics that measure local accuracy, continuity, and global accuracy, respectively. 

The contributions of this work are three-fold. First, we propose contextual text detection, a new text detection setup that aims to detect contextual text blocks that transmit complete text messages. To the best of our knowledge, this is the first work that studies the contextual text detection problem. Second, we design CUTE, a contextual text detector that detects integral text units and groups them into contextual text blocks in natural reading order. Third, we create two well-annotated datasets on contextual text detection and introduce three metrics to evaluate contextual text detection from multiple perspectives.

\section{Related Works}
\subsection{Scene Text Detection}
Recent scene text detectors can be broadly classified into two categories. The first category takes a bottom-up approach which first detects low-level text elements and then groups them into words or text lines. For example, CRAFT~\cite{baek2019character} and SegLink~\cite{Shi_2017_CVPR,tang2019seglink++}  detect characters or small segments of text instance and link them together to form text bounding boxes. The second category treats words as one specific type of objects and detects them directly by adapting various generic object detection techniques. For example, EAST \cite{Zhou_2017_CVPR}, TextBoxes++ \cite{liao2018textboxes++}, RRD \cite{liao2018rotation} and PSENet \cite{wang2019shape} detect text bounding boxes directly with generic object detection or segmentation techniques. Recent studies further improve by introducing border or counter awareness \cite{xue2018accurate,wang2020contournet,zhu2021fourier,dai2021progressive}, local refinement \cite{zhang2019look,he2021most}, deformation convolution \cite{wang2018geometry,xiao2020sequential}, Bezier curve \cite{liu2020abcnet}, etc. Besides, document layout analysis \cite{clausner2017icdar2017,zhong2019publaynet,jaume2019funsd,michael2021icpr,long2022towards} have been studied for years that usually take reading order of texts in document as consideration.

The existing scene text detectors have achieves very impressive performance. However, they are designed to detect individual text units like characters or words while the contextual information is missed. Differently, we propose a new setup that aims to detect contextual text blocks that deliver complete text messages.

\subsection{Sequence Modeling}
Sequence modeling has been widely studied in the field of NLP. Seq2Seq \cite{sutskever2014sequence} presents an encoder-decoder structure for sequential natural language processing by using Recurrent Neural Network (RNN) \cite{rumelhart1985learning}. Attention mechanisms \cite{bahdanau2014neural,luong2015effective} is also introduced to relate different positions of a single sequence in order to compute a representation of the sequence. More recently, the advanced Transformer \cite{vaswani2017attention} is proposed which relies entirely on self-attention to compute representations of the input and output without using sequence-aligned RNNs or convolution.

Sequence modeling has also been adopted in computer vision tasks. RNNs \cite{shi2016end,su2014accurate} and Transformers \cite{Yu_2020_CVPR,xue2021i2c2w,xu2020layoutlm} have been widely used in recent scene text recognition studies since most scene texts are sequentially placed in scenes. Some work also studies visual permutation for Jigsaw puzzle \cite{santa2017deeppermnet,noroozi2016unsupervised}. With the recent advances in Transformers, some work models different computer vision tasks sequentially in image recognition \cite{dosovitskiy2020image}, object detection \cite{carion2020end}, etc. More recently, \cite{li2020end,wang2021general} learn text sequences in document analysis by using Graph Convolution Network (GCN) \cite{kipf2017semi}.

We propose a contextual text detector which detects integral texts and groups them into contextual text blocks by attention mechanism. Different from existing work, the proposed CUTE can detect multiple contextual text blocks that convey different text messages in one image.

\section{Problem Definition}
In this section, we formalize the definition of terminologies in the contextual text detection problem. 

\noindent\textbf{Integral Text Unit:} We define the basic detection units as integral text units which are usually integral components of a contextual text block. These units could be characters, words to text lines, depending on different real-world scenarios and applications. In contextual text detection problem, each integral text unit in image $\mI \in \sR^{3 \times H \times W}$ is localized by using a bounding box $\vt$ by:
\begin{equation}
\begin{gathered}
\vt = (\vp_0, \vp_1, ..., \vp_{k-1}),\\
\vp_i = (x_i, y_i), x_i \in [0, W-1], y_i \in [0, H-1], 
\end{gathered}
\label{eq:integral}
\end{equation}
where $k$ is the number of vertices in bounding boxes and it varies depending on different shapes of bounding boxes.

\noindent\textbf{Contextual Text Block:} A contextual text block is defined by a set of integral text units arranged in natural reading order. It delivers a complete text message which can be one or multiple sentences lying in one or multiple lines. Each contextual text block $\vc$ is defined by:
\begin{equation}
\vc = (\vt_0, \vt_1, ..., \vt_{m-1}),
\label{eq:contextual}
\end{equation}
where $m$ is the number of integral text units in $C$. 

\noindent\textbf{Contextual Text Detection:} Given an input image $\mI \in \sR^{3 \times H \times W}$, contextual text detection aims for a model $\displaystyle f$ that can predict a set of contextual text blocks by :
\begin{equation}
\mC = \displaystyle f(\mI), \:\:\:\:\: \mC = \{\vc_0, \vc_1, ..., \vc_{n-1}\},
\end{equation}
where $n$ is the number of contextual text blocks in $\mI$.

\begin{figure}[!t]
  \centering
  \includegraphics[width=\linewidth]{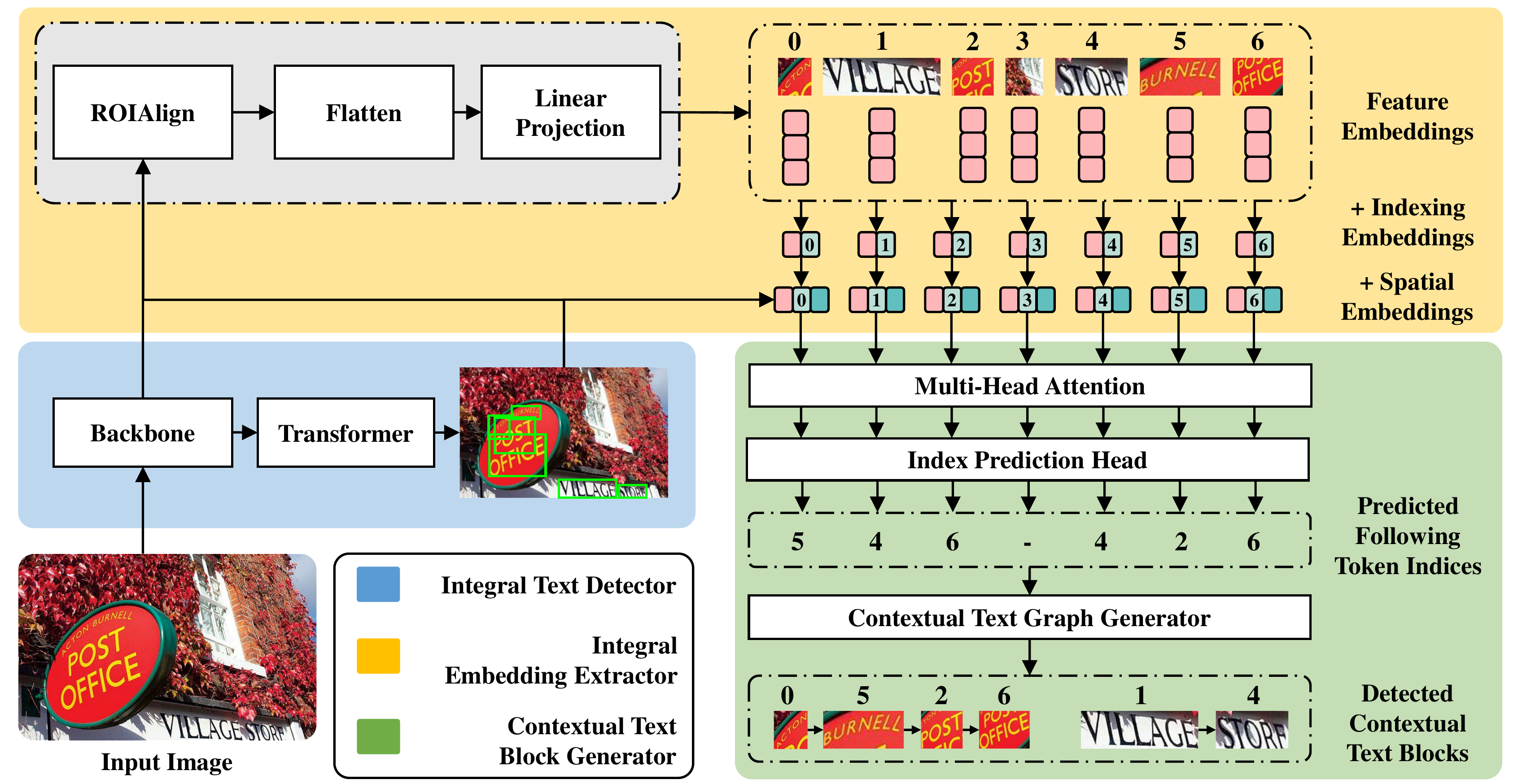}
\caption{\textbf{The framework of the proposed contextual text detector (CUTE):} Given a scene text image as input, CUTE first detects integral text units with an \textit{Integral Text Detector}. For each detected integral text unit, it then learns textual \textit{Feature Embeddings}, \textit{Indexing Embeddings} and \textit{Spatial Embeddings} that capture visual text features, text order features, and text spatial adjacency features, respectively. Finally, it models the relationship of integral text units by learning from the three types of embeddings with a \textit{Contextual Text Block Generator} and produces contextual text blocks that convey complete text messages.
}

\label{fig:network}
\end{figure}

\section{Method}

We propose a network CUTE for contextual text detection which consists an \textit{Integral Text Detector}, an \textit{Integral Embedding Extractor} and a \textit{Contextual Text Block Generator} as illustrated in Fig. \ref{fig:network}. The \textit{Integral Text Detector} first localizes a set of integral text units from input images. The \textit{Integral Embedding Extractor} hence learns visual and contextual feature embeddings for each detected integral text unit. Finally, the \textit{Contextual Text Block Generator} groups and arranges the detected integral texts in reading order to produce contextual text blocks.

\subsection{Integral Text Detector}
We adopt Transformer-based generic object detector \cite{carion2020end} as the integral text detector in our CUTE which is built upon CNN and Transformer architecture. Given an input image $\mI \in \sR^{3 \times H \times W}$, the DETR first extracts image features $\vx \in \sR^{3 \times H_0 \times W_0}$ by using a CNN backbone (e.g., ResNet \cite{he2016deep}). A Transformer hence predicts bounding boxes $\vt$ (in Equation \ref{eq:integral}) of integral text units from the extracted features $\vx$. 

One of the major advances of the Transformer-based detector is that the Transformer models all interactions between elements of image features for object detection. Specifically, the feature map $\vx$ is first flattened to a sequence of elements (i.e., pixels) accompanied with 2D positional embeddings. The Transformer hence focuses on image regions for each object by learning the relationships between each pair of elements in feature map $\vx$. As such, we adopt Transformer-based detector as the integral text detector in our CUTE for better modelling of element interactions in the visual features from network backbone. More details are available in the Supplementary Material.

\subsection{Integral Embedding Extractor}
Both visual and contextual features of integral text units are indispensable to accurate detection of contextual text blocks. We therefore design an Integral Embedding Extractor to extract three types of embeddings for each integral text unit including: (1) feature embeddings that are learnt from visual features of integral text units; (2) indexing embeddings that are encoded for integral ordering; (3) spatial embeddings that are predicted from spatial features of integral text units.

\noindent\textbf{Feature Embeddings:} We first extract visual features of integral text units and predict a set of feature embeddings. Given the image features $x$ that are extracted from backbone network, the feature embeddings of the integral text units $\vv_{fe} \in \sR^{r \times d}$ are defined by:
\begin{equation}
\begin{gathered}
\vv_{fe} = (\vv_{fe}^0, \vv_{fe}^1, ..., \vv_{fe}^{r-1}), \\
\vv_{fe}^i = \vx_c^i W + b, \:\:\:\:\: \vx_c^i = \text{flatten}(\text{ROIAlign}(\vx, \vt_i)).
\end{gathered}
\end{equation}
Specifically, we first crop the visual features $\vx_c$ for each of detected integral text units from the image features $\vx$ by using the detected integral text boxes $\vt$ from integral text detector. These features $\vx_c$ are hence flattened and linearly projected to dimension $d$ to produce feature embeddings $\vv_{fe}$, where $r$ is the number of detected integral text units in image. More details about dimension $d$ are available in Appendix.

\noindent\textbf{Indexing Embeddings:} We also introduce indexing embeddings for integral text ordering. Given a set of detected integral text units, we assign each integral text unit with an index number $i$, where $i \in [0, r-1]$ refers to the $i$-th integral text unit. Next, we adopt sinusoidal positional encoding \cite{vaswani2017attention} on these indices to produce indexing embeddings $\vv_{ie} \in \sR^{r \times d}$ by:
\begin{equation}
\begin{gathered}
\vv_{ie} = (\vv_{ie}^0, \vv_{ie}^1, ..., \vv_{ie}^{r-1}), \\
\vv_{ie}^i = \begin{cases}
                                                                      sin(i/10000^{2d_k/d}), & \text{if $d_k = 2n$},\\
                                                                      cos(i/10000^{2d_k/d}), & \text{if $d_k = 2n + 1$}.
                                                                                  \end{cases}
\end{gathered}  
\end{equation}

\noindent\textbf{Spatial Embeddings:} The spatial information of each detected integral text unit (i.e., size and position of integral texts in images) are lost because integral text features are extracted by cropping and resizing. For accurate contextual text block detection, we introduce spatial embeddings that encodes the spatial information to each integral text unit. Specifically, we use a vector $\vv_s^i$ to represent the spatial information of $i$-th integral text unit which is defined by:
\begin{equation}
\vv_s^i = (w, h, x_1, y_1, x_2, y_2, w \times h),
\end{equation}
where $w$, $h$, ($x_1$, $y_1$) and  ($x_2$, $y_2$) refer to the width, height, top-left vertex coordinate, and bottom-right vertex coordinate of integral text bounding box $t^i$. The spatial embeddings $\vv_{se} \in \sR^{r \times d}$ are hence obtained by two linear transformations:
\begin{equation}
\begin{gathered}
\vv_{se} = (\vv_{se}^0, \vv_{se}^1, ..., \vv_{se}^{r-1}), \\
\vv_{se}^i =\text{max}(0, \text{max}(0, \vv_{s}^i W_1 + b_1)W_2 + b_2).
\end{gathered}
\end{equation}
The text tokens are hence obtained by:
\begin{equation}
\vv_{token} = \text{Concat}(\vv_{fe}, \vv_{ie}, \vv_{se}).
\end{equation}

\subsection{Contextual Text Block Generator} \label{sec:ctbg}

Taking the integral tokens $\vv_{token}$ as input, the Contextual Text Block Generator groups and arranges these integral tokens in reading order. As illustrated in Fig. \ref{fig:network}, it learns the relationship between each pairs of integral tokens $\vv_{token}$ by a multi-head attention layer and produces contextual text blocks by an index prediction head and a contextual text graph generator.

\noindent\textbf{Multi-Head Attention:} We use multi-head self-attention mechanism to model the relationships between each pair of integral text units. Six stacked attention modules are adopted and each of them contains a multi-head self-attention layer following by a linear transformation layer. Layer normalization \cite{ba2016layer} is applied to the input of each layer. The text tokens $\vv_{token}$ serve as values, keys, and queries of the attention function.

\noindent\textbf{Index Prediction Head:} We model the contextual information learning as an index classification problem by an index prediction head. We adopt a linear projection layer is predict a set of indices $\textbf{\text{INX}} = (\text{INX}^0, \text{INX}^1, ...,  \text{INX}^{r-1})$, where $\vv_{token}^{j}$ follows $\vv_{token}^i$ in reading order if $\text{INX}^i = j$. Cross-entropy loss is adopted for network optimization. Note we assign a random but unique index to each detected integral text unit as the detected integral text units are usually in random order.

For the $i$-th indexed query token $\vv_{token}^i$, three cases are considered including: (1) if $\vv_{token}^i$ is not the last integral text in a contextual block, the index prediction head outputs index class $j$ if $\vv_{token}^j$ follows $\vv_{token}^i$; (2) if $\vv_{token}^i$ is the last integral unit in a contextual block, the class $i$ will be predicted; (3) if $v_{token}^i$ is not a text (i.e., false alarms from Integral Text Detector), it will be classified to `not a text' class. In this way, a $(N+1)$-way classification problem is defined for index prediction where `$N$' refers the number of index categories and `$+1$' is the `not a text' category. `$N$' is a fixed number that is significantly larger than the possible number of integral text units in an image.

\noindent\textbf{Contextual Text Graph Generator:} A directed contextual text graph $G = (V, E)$ is constructed by a Contextual Text Graph Generator which considers the detected integral text units as vertices $V$. The $E$ refers to the edges of the graph $G$ that is obtained from the Index Prediction Head (IPH) by $E = \{(V_i, V_j) | \text{IPH}(\vv_{token}^i) = j, i \in |V|, j \in |V|\}$. A set of weakly connected components $G' = \{G'_0, G'_1, ... G'_n\}$ are produced from graph $G$ 
where $n$ refers to the number of contextual text blocks in the image. Each $G'_i = (V'_i, E'_i)$ represents a contextual text block in image where $V'_i$ refers its integral text units and $E'_i$ produces their reading order.

\begin{table*}[!t]
\centering
\caption{The \textbf{statistics} of the ReCTS-Context and SCUT-CTW-Context datasets: `integral': Integral Text Units; `block': Contextual Text Blocks; `\#': Number.}
\resizebox{\columnwidth}{!}{
\begin{tabular}{cccccccc}
\toprule
Dataset & \begin{tabular}[c]{@{}c@{}}Integral\\ Text\end{tabular} & \# integral & \# block & \# image & \begin{tabular}[c]{@{}c@{}}\# integral\\ per block\end{tabular} & \begin{tabular}[c]{@{}c@{}}\# integral\\ per image\end{tabular} & \begin{tabular}[c]{@{}c@{}}\# block\\ per image\end{tabular} \\ 
\midrule
ReCTS-Context                       & Character       & 440,027  & 107,754 & 20,000 & 4.08    & 22.00    & 5.39    \\ 
\midrule
SCUT-CTW-Context                    & Word         & 25,208   & 4,512   & 1,438  & 5.56    & 17.65    & 3.17    \\ 
\bottomrule
\end{tabular}
}
\label{tab:statistic}
\end{table*}

\section{Datasets and Evaluation Metrics}
\subsection{Datasets}
\begin{figure}[!t]
  \centering
  \includegraphics[width=\linewidth]{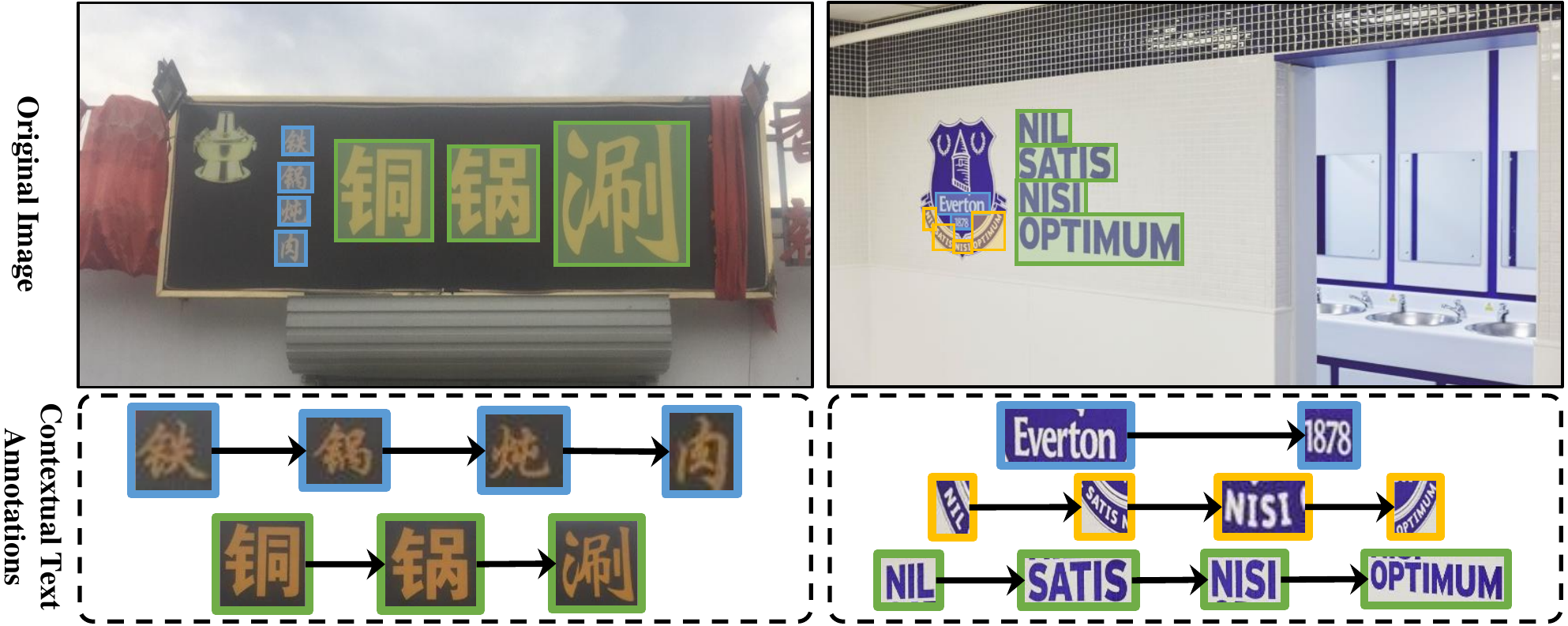}
\caption{\textbf{Illustration of contextual text block annotation:} We annotate each contextual text block by an ordered sequence of integral text units (characters or words) which together convey a complete textual message. The two sample images are picked from datasets ReCTS and SCUT-CTW, respectively.}
\label{fig:dataset_samples}
\end{figure}

We create two contextual-text-block datasets ReCTS-Context and SCUT-CTW-Context as shown in Table \ref{tab:statistic}. Fig. \ref{fig:dataset_samples} shows two samples where integral text units belonging to the same contextual text block are grouped in proper order. 

\noindent\textbf{ReCTS-Context (ReCTS):} We annotate contextual text blocks for images in dataset ICDAR2019-ReCTS~\cite{zhang2019icdar}, which are split into a training set and a test set with 15,000 and 5,000 images, respectively. It contains largely Chinese texts with characters as integral text units.

\noindent\textbf{SCUT-CTW-Context (SCUT-CTW):} We annotate contextual text blocks for dataset SCUT-CTW-1500 dataset~\cite{yuliang2017detecting} which consists of 940 training images and 498 test images. Most integral text units in this dataset are words which have rich contextual information as captured in various scenes. More details about the two created datasets are available in Appendix.

\subsection{Evaluation Metrics}
We propose three evaluation metrics for the evaluation of contextual text detection:

\noindent\textbf{Local Accuracy (LA):} We introduce LA to evaluate the accuracy of order prediction for neighbouring integral text units. Considering two correctly detected integral text units $a$ and $b$ (with $b$ following $a$ as ground-truth), a true positive is counted if the detection box of $b$ is predicted as directly following that of $a$. We compute LA by $LA = TP / N$ where $TP$ denotes the number of true positives and $N$ is the total number of connected pairs in ground-truth.

\addtolength{\tabcolsep}{2pt}
\begin{table*}[t]
\centering
\caption{Quantitative comparison of CUTE with state-of-the-art methods on \textbf{ReCTS-Context}. LA: Local Accuracy; LC: Local Continuity; GA: Global Accuracy.}
\begin{tabular}{lccccccccc}
\toprule
\multirow{2}{*}{Model} & \multicolumn{3}{c}{IoU=0.5} & \multicolumn{3}{c}{IoU=0.75} & \multicolumn{3}{c}{IoU=0.5:0.05:0.95} \\ \cmidrule(lr){2-4} \cmidrule(lr){5-7} \cmidrule(lr){8-10}
                       & LA       & LC      & GA      & LA       & LC       & GA      & LA          & LC          & GA         \\ \midrule
CLUSTERING \cite{cheng1995mean}             & 32.22   & 19.06  & 10.59  & 26.06   & 17.01   & 9.66  &  25.60     & 16.13      & 9.02     \\ 
\midrule
CRAFT-R50 \cite{baek2019character}          & 63.66   & 53.26  & 45.96  & 51.22   & 48.39   & 36.76  &  50.06     & 45.46      & 35.60     \\
LINK-R50 \cite{xue2021detection}            & 68.15   & 57.50  & 48.39  & 53.83   & 50.19   & 38.36  &  52.95     & 47.69     & 37.33     \\
\textbf{CUTE-R50}                            & \textbf{70.43}   & \textbf{64.74}  & \textbf{51.55}  & \textbf{54.39}   & \textbf{56.63}   & \textbf{39.52}  & \textbf{53.92} &  \textbf{53.56}   &  \textbf{38.92}      \\ 
\midrule
CRAFT-R101  \cite{baek2019character}        & 65.21    & 54.59 & 47.02   & 52.01    & 48.65  & 37.21   & 51.56      &   46.10        &  36.33     \\
LINK-R101   \cite{xue2021detection}         & 70.78    & 59.10    & 49.92    &  54.53  & 51.02  & 38.98     &  53.42      & 48.26     &  37.94      \\
\textbf{CUTE-R101}                           & \textbf{72.36}  & \textbf{67.33}   &   \textbf{53.76}  &  \textbf{55.14}  &  \textbf{57.03}   & \textbf{40.21}   &  \textbf{54.56}   &  \textbf{53.94}           &  \textbf{39.42}          \\ 
\bottomrule
\end{tabular}
\label{tab:result_rects}
\end{table*}
\addtolength{\tabcolsep}{-2pt} 

\addtolength{\tabcolsep}{2pt}
\begin{table*}[t]
\centering
\caption{Quantitative comparison of CUTE with state-of-the-art methods on \textbf{SCUT-CTW-Context}. LA: Local Accuracy; LC: Local Continuity; GA: Global Accuracy.}
\begin{tabular}{lccccccccc}
\toprule
\multirow{2}{*}{Model} & \multicolumn{3}{c}{IoU=0.5} & \multicolumn{3}{c}{IoU=0.75} & \multicolumn{3}{c}{IoU=0.5:0.05:0.95} \\ \cmidrule(lr){2-4} \cmidrule(lr){5-7} \cmidrule(lr){8-10} 
                       & LA       & LC      & GA      & LA       & LC       & GA      & LA          & LC          & GA         \\ \midrule
CLUSTERING \cite{cheng1995mean} & 18.36   & 7.93  & 6.78  & 14.11   & 5.88   & 4.72  & 13.54      & 5.71      & 4.88     \\ 
\midrule
LINK-R50 \cite{xue2021detection}& 25.47   & 3.33  & 18.88  & 20.25   & 3.15   & 14.70  &  19.31     & 2.93     & 14.26     \\
\textbf{CUTE-R50}               & \textbf{54.01}   & \textbf{39.19}  & \textbf{30.65} & \textbf{41.62}   & \textbf{31.19}   & \textbf{23.71}  & \textbf{39.44}      & \textbf{29.03}      & \textbf{22.10}     \\
\midrule
LINK-R101 \cite{xue2021detection}& 25.71   & 3.41  & 19.18  & 20.02   & 2.89   & 14.68  &  19.56     & 2.72     & 14.39     \\
\textbf{CUTE-R101}              & \textbf{55.71}   & \textbf{39.38}  & \textbf{32.62}  & \textbf{40.61}   & \textbf{29.04}   & \textbf{22.77}  & \textbf{39.95}      & \textbf{28.30}      & \textbf{22.69}     \\ 
\bottomrule
\end{tabular}
\label{tab:result_ctw}
\end{table*}
\addtolength{\tabcolsep}{-2pt}

\noindent\textbf{Local Continuity (LC):} We introduce LC to evaluate the continuity of integral text units by computing a modified $n$-gram precision score as inspired by BLEU \cite{papineni2002bleu}. Specifically, we compare $n$-grams of the predicted consecutive integral text units with the $n$-grams of the ground-truth integral texts and count the number of matches, where $n$ varies from 1 to 5. For $n = 1$, we only consider the scenario that the contextual text block contains one integral text. 

\noindent\textbf{Global Accuracy (GA):} Besides LA and LC which focus on local characteristics of integral text units ordering, we also evaluate the detection accuracy of contextual text blocks. $TP$ is counted if all integral texts in a contextual text block are detected and the reading orders are accurately predicted. The global accuracy is hence computed by $GA = TP / N$ where $N$ is the total number of contextual text blocks in ground-truth.

Besides, a detected integral text unit is determined to be matched with ground-truth text if the intersection-over-union (IoU) of these two bounding boxes are larger than a threshold. We adopt three IoU threshold standards that are widely-used in generic object detection task \cite{liu2020deep} including $IoU = 0.5, IoU = 0.75$ and $IoU = 0.5:0.05:0.95$ for thorough evaluation.

\section{Experiments}

\subsection{Comparing with State-of-the-art}

\begin{figure}[t]
  \centering
  \includegraphics[width=\linewidth]{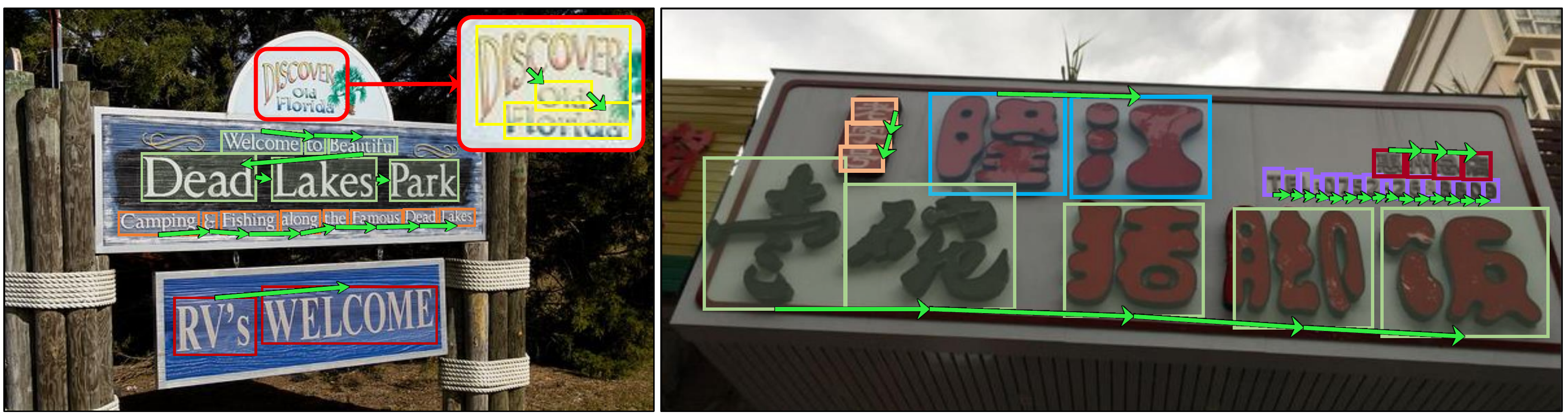}
\caption{\textbf{Illustration of the proposed CUTE:} Sample images are collected from SCUT-CTW-Context and ReCTS-Context datasets, where the color boxes highlight the detected integral text units and the green arrows show the predicted text orders. The integral text units of each contextual text block are highlighted in the same color.
}
\label{fig:sample_result}
\end{figure}

We evaluate the proposed CUTE on ReCTS-Context and SCUT-CTW-Context datasets qualitatively and quantitatively as shown in Fig. \ref{fig:sample_result} and Table \ref{tab:result_rects}-\ref{tab:upper_bound}.

Since there is little prior research on contextual text block detection, we develop a few baselines for comparisons. The first baseline is CLUSTERING that groups integral text units by mean shift clustering \cite{cheng1995mean}. The second and the third baselines are CRAFT \cite{baek2019character} and LINK \cite{xue2021detection}, two bottom-up scene text detection methods that group characters/words to text lines. Since both CRAFT and LINK do not have the concept of contextual text blocks, we sort integral text units within each contextual text block according to the common reading order of left-to-right and top-to-down. In addition, we evaluate with two backbones ResNet-50 and ResNet-101 (denoted by `R50’ and `R101') to study the robustness of the proposed CUTE.

\addtolength{\tabcolsep}{2pt}
\begin{table}[t]
\centering
\caption{Quantitative comparison of CUTE with state-of-the-art methods on \textbf{integral text grouping and ordering task}: The ground-truth integral text bounding boxes are adopted for evaluations on integral text grouping and ordering task only. LA: Local Accuracy; LC: Local Continuity; GA: Global Accuracy.} 
\begin{tabular}{lcccccc}
\toprule
\multirow{2}{*}{Model} & \multicolumn{3}{c}{SCUT-CTW} & \multicolumn{3}{c}{ReCTS} \\ \cmidrule(lr){2-4} \cmidrule(lr){5-7} 
                       & LA      & LC      & GA     & LA       & LC       & GA      \\ \midrule
CLUSTERING \cite{cheng1995mean} & 27.94   & 12.74   & 10.76  & 69.70     & 49.15    & 32.20    \\ 
\midrule
LINK-R50 \cite{xue2021detection} & 30.17 & 4.48  &  22.84  & 83.77  & 68.44  & 61.10 \\
\textbf{CUTE-R50}               & \textbf{71.48}   & \textbf{58.53}   & \textbf{49.67}  & \textbf{92.08}    & \textbf{82.79}     & \textbf{76.02}    \\
\midrule
LINK-R101 \cite{xue2021detection} & 45.54 & 6.28  &  31.69 &  86.66 & 75.03  & 69.55 \\
\textbf{CUTE-R101}              & \textbf{71.54}   & \textbf{58.68}   & \textbf{52.57}  & \textbf{93.12}    & \textbf{83.70}     & \textbf{77.81}        \\ 
\bottomrule
\end{tabular}
\label{tab:upper_bound}
\end{table}
\addtolength{\tabcolsep}{-2pt}

We compare CUTE with the three baselines over ReCTS where integral text units are at character level. As Table \ref{tab:result_rects} shows, the clustering-based method cannot solve the contextual text detection problem effectively because the integral text units are usually with different sizes, positions, and orientations. The bottom-up scene text detectors work better by focusing on visual features only. The proposed CUTE performs the best consistently as it models the relation between each pair of integral text units by considering both visual representative features and contextual information.

We further conduct experiments over SCUT-CTW where integral text units are at word level. We compare CUTE with CLUSTERING and LINK only because CRAFT cannot group texts lying on different lines. As Table \ref{tab:result_ctw} shows, CLUSTERING achieves very low performance due to the complex contextual relations among integral text units. LINK obtains very low scores on LC, showing that only short contextual text blocks with small number of integral text units are detected. CUTE instead outperforms all three baselines by large margins consistently across LA, LC and GA. Note the detection performances over SCUT-CTW are relatively low because it contains many texts with more complex layouts as compared with ReCTS.

Additionally, to validate CUTE’s effectiveness on the grouping and ordering of integral text units alone, we assume that all integral text units are accurately detected by feeding the bounding boxes of ground-truth integral text units to the \textit{Integral Embedding Extractor} and \textit{Contextual Text Block Generator}. As Table \ref{tab:upper_bound} shows, the proposed CUTE groups and orders integral text units effectively.

\begin{figure}[!t]
  \centering
  \includegraphics[width=\linewidth]{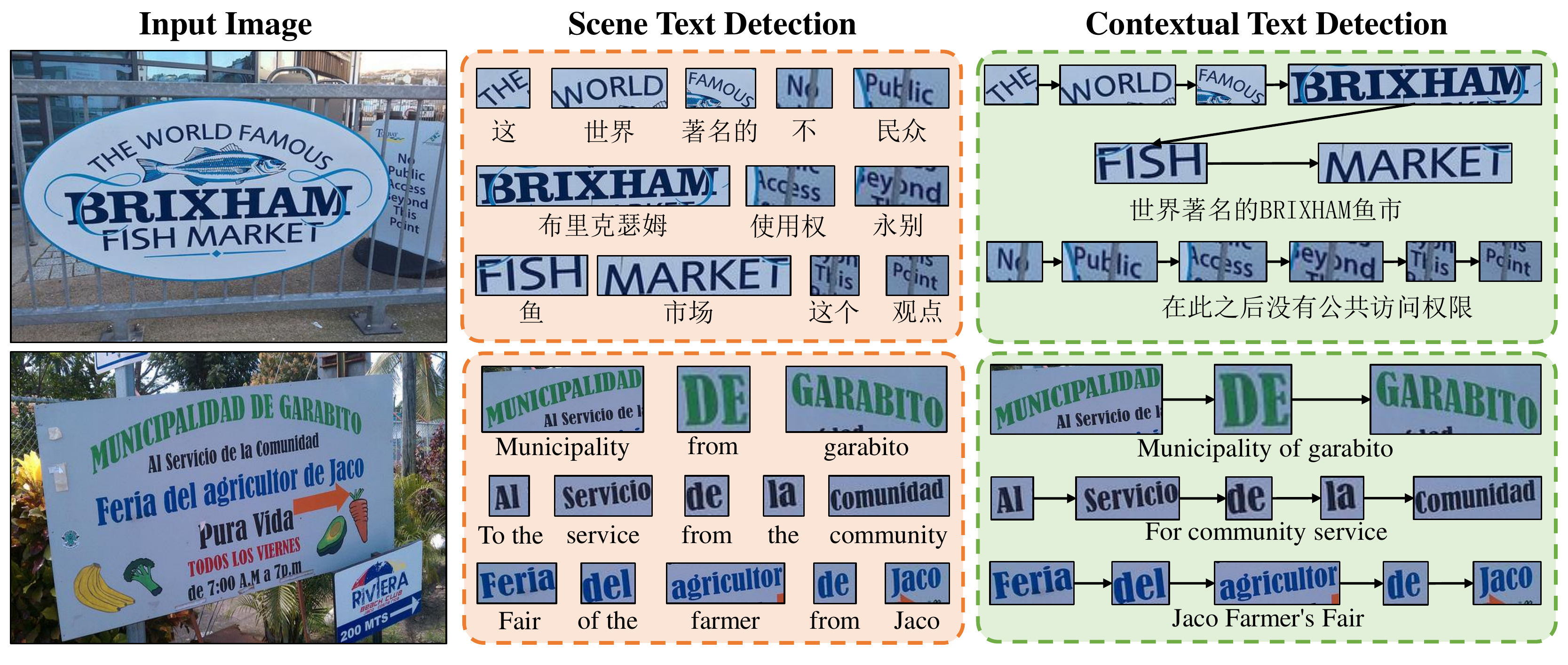}
\caption{\textbf{Contextual text detection facilitates scene text translation significantly:} The output of CUTE conveys complete text messages which can be better translated to other languages as `natural language’ with rich contextual information as shown in column 3. As a comparison, scene text detectors produce individual text units which can not be translated well as shown in column 2.}
\label{fig:translate_appendix}
\end{figure}

\addtolength{\tabcolsep}{5pt}
\begin{table}[t]
\centering
\caption{\textbf{Ablation studies} of CUTE over SCUT-CTW. LA: Local Accuracy; LC: Local Continuity; GA: Global Accuracy.}
\begin{tabular}{ccccccc}
\toprule
Model & $\vv_{fe}$ & $\vv_{se}$ & $\vv_{ie}$  & LA & LC & GA \\ 
\midrule
1     & \checkmark     &                &               & 6.86  & 3.34 & 1.94    \\ 
2     & \checkmark     & \checkmark         &               & 8.99  & 4.56 & 2.18    \\ 
3     & \checkmark     &                & \checkmark        & 28.65  & 25.71 & 21.89    \\ 
4     & \checkmark     & \checkmark         & \checkmark        & \textbf{71.48}  & \textbf{58.53} & \textbf{49.67}    \\ 
\bottomrule
\end{tabular}
\label{tab:ablation}
\end{table}
\addtolength{\tabcolsep}{-5pt}

\subsection{Ablation Study}
The proposed CUTE detects contextual text blocks by using both visual features (feature embeddings) and contextual features that capture spatial and ordering information, respectively. We conduct ablation studies over SCUT-CTW-Context to identify the contribution of each embedding. We trained four models with different combinations of the three embeddings. As Table \ref{tab:ablation} shows, CUTE does not work well with either feature embeddings alone or feature embeddings plus spatial embeddings. However, combining feature embeddings with indexing embeddings improves detection greatly as indexing embeddings introduce crucial text order information. The combination of all three embeddings performs the best by large margins, demonstrating the complementary nature of the three embeddings. 

\addtolength{\tabcolsep}{4pt}
\begin{table}[t]
\centering
\caption{The significance of contextual text detection to \textbf{scene text detection task}: The proposed CUTE effectively helps to improve scene text detection performance of different detectors (in mAP) by filtering out the false alarms.}
\begin{tabular}{ccc}
\toprule
Model                       & w/o CUTE & with CUTE \\
\midrule
PSENet\cite{wang2019shape}  & 52.30    & 53.69 (+1.39)     \\
MSR\cite{xue2019msr}        & 60.07    & 61.80 (+1.73)    \\
DETR\cite{carion2020end}    & 56.11    & 57.37 (+1.26)    \\
LINK\cite{xue2021detection} & 62.03    & 62.84 (+0.81)\\
\bottomrule
\end{tabular}
\label{tab:detection}
\end{table}
\addtolength{\tabcolsep}{-4pt}

\addtolength{\tabcolsep}{4pt}
\begin{table}[t]
\centering
\caption{The significance of contextual text detection to \textbf{text classification task:} The proposed CUTE effectively helps to improve text classification performance of different text classifiers by learning from recognized texts in contextual text blocks.}
\begin{tabular}{ccc}
\toprule
Model                       & w/o CUTE & with CUTE \\
\midrule
TextCNN\cite{kim2014convolutional}  & 90.56     & 92.40 (+1.84)     \\
TextRNN\cite{liu2016recurrent}        & 79.55    & 87.20 (+7.65)    \\
Fast Text\cite{bojanowski2017enriching}    & 90.96    & 91.82 (+0.86)    \\
Transformer\cite{vaswani2017attention} & 89.69    & 92.54 (+2.85)\\
\bottomrule
\end{tabular}
\label{tab:nlp}
\end{table}
\addtolength{\tabcolsep}{-4pt}

\subsection{Discussion}

\noindent\textbf{Contextual text detection facilitates downstream tasks:} The proposed detection setup for contextual text blocks can facilitate both scene text detection and many downstream tasks. We first study how the proposed contextual text detection can improve the scene text detection task over SCUT-CTW. Specifically, traditional scene text detectors tend to produce false detection at image background that has similar visual representations as scene texts. CUTE can suppress such false detection effectively (i.e., classify the false detection into `not a text' category) by learning the text ordering through not only visual features but also contextual information of texts (details in Section \ref{sec:ctbg}). As Table \ref{tab:detection} shows, CUTE improves the scene text detection performance consistently across a number of scene text detectors that adopt different backbones and detection strategies.

We also study how the proposed contextual text detection can facilitate various downstream tasks. We focus on the scene text translation task that is very useful to scene understanding for visitors with different home languages. Specifically, we feed each detected text (i.e. a character, word, or contextual text block) to a neural machine translator (Google Translator) for translation across different languages. As Fig. \ref{fig:translate_appendix} shows, CUTE groups and orders scene texts into contextual text blocks (delivering complete textual messages) which facilitates scene text translation greatly as compared with traditional scene text detectors without the concept of contextual text blocks. 

We additionally study how contextual text detection facilitates text classification task in NLP. Specifically, we classify and annotate the transcription of texts in ReCTS-Context into three categories (i.e., `Address', `Phone Number' or `Restaurant Name') according to the text semantics. We train models by different text classification techniques and test on the detected texts from integral texts (denoted by `w/o CUTE') and contextual text blocks (denoted by `with CUTE'), respectively. As shown in Table \ref{tab:nlp}, the use of CUTE helps to improve the text classification consistently across different text classifiers.

\begin{figure}[!t]
  \centering
  \includegraphics[width=\linewidth]{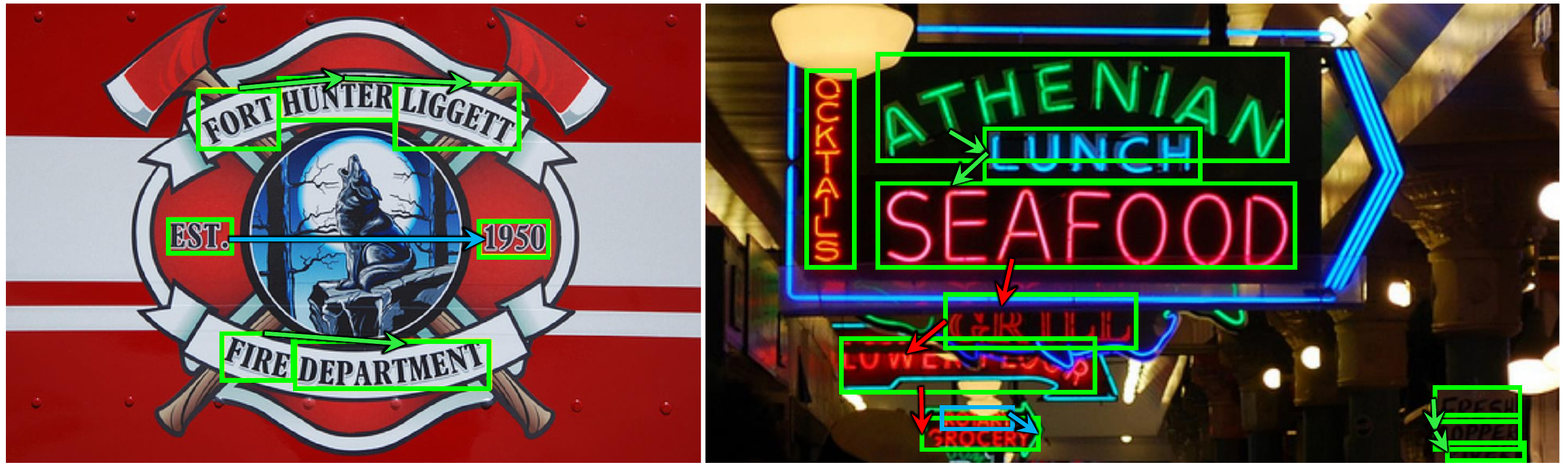}
\caption{\textbf{Typical failure cases of the proposed CUTE:} Correct, incorrect and missing orders or integral text units are highlighted by arrows in yellow, red and blue, respectively. The proposed CUTE may fail if the images contain complex text layouts.}
\label{fig:failure}
\end{figure}

\noindent\textbf{Typical failure cases:} The proposed CUTE may fail if the images contain complex text layouts. As shown in Fig. \ref{fig:failure}, the proposed CUTE may fail if the texts from different contextual text blocks are in similar font styles and extremely close to (or far from) each other.

\section{Conclusion and Future Work}\label{sec:future}
We study contextual text detection, a new text detection setup that first detects integral text units and then groups them into contextual text blocks. We design CUTE, a novel method that detects contextual text blocks effectively by combining both visual and contextual features. In addition, we create two contextual text detection datasets within which each contextual text block is well annotated by an ordered text sequence. Extensive experiments show that CUTE achieves superior contextual text detection, and it also improves scene text detection and many downstream tasks significantly. 

In the future, we will continue to study contextual text detection when scene texts have complex layouts. Specifically, we will expand and balance our datasets by including more complex scenes and text layouts. We will also study how to leverage text semantics (from scene text recognition) for better contextual text detection.

\bibliographystyle{splncs04}
\bibliography{egbib}
\end{document}